\documentclass{article}
\usepackage{spconf,amsmath,graphicx,hyperref}
\usepackage{algorithm}
\usepackage{algpseudocode}
\usepackage{booktabs}
\usepackage{graphicx}
\usepackage{enumitem}
\usepackage{url}
\usepackage{hyperref}
\usepackage{capt-of}
\usepackage{xcolor}

\title{Scalable Subgraph Sampling via Resistance Curvature}
\name{Chaoqun~Fei$^{1}$, Tinglve~Zhou$^{1}$, Tianyong~Hao$^{2}$, Yangyang~Li$^{3,*}$\thanks{$^{*}$Corresponding author: Yangyang Li.}}
\address{
$^{1}$ School of Artificial Intelligence, South China Normal University\\ 
$^{2}$ School of Computer Science, South China Normal University\\
$^{3}$ Academy of Mathematics and Systems Science, Chinese Academy of Sciences
}

\begin{document}

\maketitle

\begin{abstract}
Subgraph sampling reduces the training cost of large-scale graph neural networks, but sampling criteria may overlook the geometric roles of edges. We propose a resistance-curvature-guided sampling framework built on ERC-LG, a curvature approximation method for large-scale graphs. ERC-LG combines Johnson-Lindenstrauss projections with regularized multi-GPU batched conjugate gradient solvers, avoiding explicit Laplacian pseudoinverse computation and full embedding storage. The resulting curvature informs node- and edge-sampling probabilities for constructing GNN training subgraphs. Experiments show numerical agreement with pseudoinverse-based curvature and reduced runtime compared with CG-only computation. ERC-LG-based sampling variants achieve the highest mean accuracy on six of seven real-world datasets in downstream node classification.

\end{abstract}
\begin{keywords}
Effective Resistance Curvature, JL Projection, batched CG, Subgraph sampling
\end{keywords}
\vspace{-10pt}
\section{Introduction}
\vspace{-5pt}
\label{sec:intro}

Graph neural networks (GNNs) face substantial memory and computational costs when trained on large graphs. Neighborhood~\cite{hamilton2017graphsage}, layer-wise~\cite{chen2018fastgcn}, and subgraph~\cite{chiang2019clustergcn} sampling provide scalable alternatives to full-graph training. However, sampling may discard connections important for information propagation, motivating sampling criteria that account for the structural roles of edges.

Discrete graph curvature provides edge-level geometric information for distinguishing redundant connections from potential bottlenecks and bridges. ORG-sub~\cite{wu2023subsampling} uses Ollivier-Ricci curvature to guide sampling and improve small-community coverage. However, it focuses on community analysis rather than GNN training, and its scalability on massive graphs remains unverified. Moreover, exact Ollivier-Ricci curvature requires edge-wise optimal transport computations, making graph-wide evaluation expensive.

Our previous work introduced an efficient effective resistance curvature computation scheme\cite{fei2025efficientcurvatureawaregraphnetwork}. Effective resistance curvature (ERC) uses effective resistance to capture multipath connectivity without optimal transport. By replacing Laplacian pseudoinversion with diagonally perturbed matrix inversion, the scheme achieved speedups of up to approximately $1{,}000\times$. It maintained near-identical curvature values to the pseudoinverse-based reference and comparable performance to Ollivier-Ricci curvature in structural discrimination and graph representation learning. Nevertheless, explicit matrix inversion retains quadratic storage requirements, limiting its applicability to larger graphs.

To address this bottleneck, we propose an efficient approximation method for large-scale graphs short for ERC-LG. The key idea is to use Johnson--Lindenstrauss (JL)
projections to reduce the number of linear systems required for resistance estimation.
Regularized multi-GPU batched CG avoids explicit pseudoinversion, while streaming distance accumulation avoids storing the full embedding. We further design ERC-LG-based sampling methods for GNN training.

Our main contributions are as follows:
\begin{enumerate}[label=\arabic*), itemsep=-4pt, topsep=0pt]
    \item We develop ERC-LG by adapting JL-based effective resistance approximation to ERC computation, together with regularized multi-GPU batched CG and streaming edge-distance accumulation, avoiding explicit pseudoinverse computation and full embedding storage.
    \item We design ERC-LG-based node and edge sampling methods that incorporate edge-level geometric information into sampling probabilities for GNN training-subgraph construction.
    \item We evaluate numerical agreement, computational efficiency, and downstream node classification. ERC-LG-based sampling variants achieve the highest mean accuracy on six of seven real-world datasets.
\end{enumerate}

\vspace{-10pt}
\section{Related Work}
\vspace{-5pt}
\subsection{Sampling for Large-Scale Graph Neural Networks}
Existing GNN sampling methods mainly include neighbor sampling, layer-wise sampling, and subgraph sampling. GraphSAGE~\cite{hamilton2017graphsage} controls computational cost by sampling a fixed number of neighbors at each layer. FastGCN~\cite{chen2018fastgcn} interprets graph convolution as an integral transform and applies layer-wise importance sampling. Subgraph sampling instead directly constructs minibatch training graphs. ClusterGCN~\cite{chiang2019clustergcn} organizes training batches through graph clustering, while GraphSAINT~\cite{zeng2020graphsaint} combines subgraph sampling with normalization corrections to reduce training cost and control sampling bias. REGNN~\cite{dingandwei2025} combines random-walk sampling and historical embeddings to enable scalable mini-batch GNN training. 
\vspace{-5pt}
\subsection{Graph Learning with Discrete Curvature}
Discrete graph curvature provides geometric information for analyzing and adapting graph structures in representation learning. Topping et al.~\cite{topping2022oversquashing} use curvature-guided rewiring to address oversquashing. RicciPool~\cite{fei2026geometricflowenhancedgraph} combines Ollivier-Ricci flow with spectral clustering for graph pooling. ORG-sub~\cite{wu2023subsampling} uses Ollivier-Ricci curvature to improve small-community coverage, while LC-sub~\cite{shu2023109475} employs a 3-cycle-based curvature approximation for combinatorial subgraph sampling. Spielman and Srivastava~\cite{spielman2011resistance} approximate effective resistance through random projections and Laplacian solves for spectral sparsification, whereas we adapt this idea to graph-wide ERC estimation and GNN sampling. Devriendt and Lambiotte~\cite{devriendt2022resistancecurvature} define node and edge resistance curvatures. Our previous work accelerates ERC computation through matrix perturbation~\cite{fei2025efficientcurvatureawaregraphnetwork}, while DGSL-RCF~\cite{fei2026dynamicgraphstructurelearning} uses resistance curvature flow for iterative graph refinement.

\vspace{-8pt}

\section{Method}
\vspace{-5pt}
\label{sec:method}
We approximate ERC using JL projections and multi-GPU batched CG, then use the resulting edge geometry to guide node and edge sampling.
\vspace{-10pt}
\subsection{ERC Approximation for Large-Scale Graphs}
\vspace{-5pt}
\subsubsection{Resistance curvature}
\vspace{-5pt}
For an undirected graph $G=(V,E)$ with $n$ nodes and $m$ edges, let $L=D-A=B^\top WB$, where $B\in{R}^{m\times n}$ is an oriented incidence matrix and $W=\operatorname{diag}(w_e)$ contains positive edge weights ($W=I$ for unweighted graphs). Let $b_{uv}=e_u-e_v$, effective resistance $R_{uv}$, node curvature $p_u$ and edge curvature $\kappa_{uv}$~\cite{devriendt2022resistancecurvature} are defined as below:
\begin{equation}
\begin{aligned}
R_{uv}&=b_{uv}^\top L^\dagger b_{uv}.
\end{aligned}
\label{eq:resistance}
\vspace{-5pt}
\end{equation}
\begin{equation}
\begin{aligned}
p_u&=1-\tfrac12\sum_{v\in\mathcal N(u)}w_{uv}R_{uv},
\quad \kappa_{uv}=\frac{2(p_u+p_v)}{R_{uv}}.
\end{aligned}
\label{eq:curvature}
\vspace{-5pt}
\end{equation}
Here, $L^\dagger$ is the Moore-Penrose pseudoinverse of $L$, and $\kappa_{uv}$ denotes the ERC of edge $(u,v)$.
\vspace{-5pt}
\subsubsection{Johnson-Lindenstrauss embedding}
\vspace{-5pt}
Utilizing $L^\dagger LL^\dagger=L^\dagger$, the effective resistance can be rewritten as a squared embedding distance:
\begin{equation}
R_{uv}=\|b_{uv}^\top L^\dagger B^\top W^{1/2}\|_2^2.
\label{eq:resistance_embedding}
\end{equation}
Following~\cite{spielman2011resistance}, we use JL projections~\cite{johnson1984extensions} to approximately preserve such pairwise distances, eliminating the need to store the full set of embedding coordinates. 
Let $Q\in{R}^{m\times K}$ have independent entries uniformly sampled from $\{\pm1/\sqrt K\}$, and define $Y=B^\top W^{1/2}Q$. Since each column of $Y$ lies in the range of $L$, the low-dimensional embedding and its associated system satisfy
\begin{equation}
Z=L^\dagger Y,
\qquad LZ=LL^\dagger Y=Y.
\label{eq:jl_system}
\vspace{-5pt}
\end{equation}
Thus, computing the embedding requires solving $K$ linear systems sharing the coefficient matrix $L$, without explicitly forming the pseudoinverse.
\vspace{-10pt}
\subsubsection{Regularization and Parallel Solution}
\vspace{-5pt}
The coefficient matrix $L$ in Eq.~\eqref{eq:jl_system} is singular. To construct a symmetric positive definite system suitable for standard CG, we follow the diagonal perturbation approach in our prior work\cite{fei2025efficientcurvatureawaregraphnetwork} and introduce $\varepsilon>0$, replacing the original system with
\begin{equation}
L_\varepsilon=L+\varepsilon I,
\qquad L_\varepsilon Z_\varepsilon=Y.
\label{eq:regularized_embedding}
\end{equation}
Since the columns of $Y$ lie in the range of $L$, $Z_\varepsilon$ approaches the target embedding $Z=L^\dagger Y$ as $\varepsilon\to0$ and finite perturbation introduces regularization bias.

We distribute the $K$ independent right-hand sides across GPUs and perform column-wise CG~\cite{hestenes1952methods} in batches on each device. Each GPU maintains a local sparse graph copy and evaluates matrix products through $L_\varepsilon X=LX+\varepsilon X$, without explicit inversion.
Let $\widehat Z_\varepsilon$ denote the approximate embedding computed by CG. Effective resistances on the original edges are estimated as
\begin{equation}
\widehat R_{uv}
=\|\widehat Z_{\varepsilon,u,:}
-\widehat Z_{\varepsilon,v,:}\|_2^2.
\label{eq:jl_resistance}
\end{equation}

After accumulating each batch's squared-distance contributions, its workspace is released, avoiding storage of the full embedding. Finally, contributions from all GPUs are aggregated, and curvature is computed using Eq.~\eqref{eq:curvature}. Algorithm~\ref{alg:scalable_rc} summarizes the procedure.

\begin{algorithm}[t]
\footnotesize
\caption{ERC Approximation for Large-scale Graphs}
\label{alg:scalable_rc}
\begin{algorithmic}[1]

\Require Graph $G$, dimension $K$, regularization
$\varepsilon$, GPUs $P$, batch size $b$,
CG tolerance $\tau$
\Ensure Edge curvatures $\widehat\kappa$

\State Construct sparse $L$ and define
$L_\varepsilon X=LX+\varepsilon X$;
partition the $K$ projection columns across $P$ GPUs

\For{GPU $g=1,\ldots,P$ \textbf{in parallel}}
    \State Store local sparse graph operators;
    $r^{(g)}\gets\mathbf{0}\in{R}^{m}$

    \For{each assigned column batch $t$ of size at most $b$}
        \State Form $Y^{(t)}=B^\top W^{1/2}Q^{(t)}$
        from streamed independent
        $Q_{ij}^{(t)}\sim
        \operatorname{Unif}\{\pm1/\sqrt K\}$

        \State Solve
        $L_\varepsilon\widehat Z_\varepsilon^{(t)}
        =Y^{(t)}$
        by batched column-wise CG to relative
        residual tolerance $\tau$

        \State Add this batch's squared edge distances
        to $r^{(g)}$ in edge blocks,
        using Eq.~\eqref{eq:jl_resistance}

        \State Release batch workspace
    \EndFor
\EndFor

\State Aggregate $\widehat R\gets\sum_g r^{(g)}$; return $\widehat\kappa$
using Eq.~\eqref{eq:curvature}
\end{algorithmic}
\end{algorithm}

\vspace{-15pt}
\subsection{Resistance-Curvature-Guided Subgraph Sampling}
\vspace{-5pt}

Node sampling and edge sampling are two basic approaches to large-graph sampling, drawing nodes or edges according to probability distributions. We use ERC-LG to construct these distributions and develop curvature-guided node and edge sampling strategies.
\vspace{-5pt}
\subsubsection{Curvature-guided Node sampling}
\vspace{-5pt}
Higher-curvature edges often occur in densely interconnected regions. We aggregate their normalized scores to favor seed nodes representing such local structures:
\begin{equation}
q_u^{\mathrm{node}}
=
\sum_{v\in\mathcal N(u)}
\frac{\widehat\kappa_{uv}-\kappa_{\min}}
{\kappa_{\max}-\kappa_{\min}+\eta},
\label{eq:node_probability}
\end{equation}
where $\kappa_{\min}$ and $\kappa_{\max}$ are the extrema of approximate edge curvature, and $\eta>0$ ensures numerical stability. By summing curvature-based scores over incident edges, the aggregation implicitly captures both node degree and local edge geometry, guiding the selection of seed nodes whose neighborhoods form training subgraphs. If all weights vanish, sampling is uniform over nonisolated nodes.
\vspace{-10pt}
\subsubsection{Curvature-guided Edge sampling}
\vspace{-5pt}
Lower-curvature edges may indicate bottlenecks or inter-region connections with limited alternative paths. To increase their retention probability, we use decreasing curvature weights:
\begin{equation}
q_{uv}^{\mathrm{edge}}
=
\sqrt{max(-\widehat\kappa_{uv}+2\kappa_{\max},0)}+1.
\label{eq:edge_probability}
\end{equation}
The square-root transformation moderates weight differences while favoring lower-curvature connections. Selected edges and their endpoints form the sampled subgraph. 

Thus, node sampling emphasizes local structural representation, while edge sampling emphasizes potentially critical connections. Sampling probabilities are obtained by normalizing these weights over all nodes or edges, respectively.

\vspace{-10pt}
\subsection{Complexity and Memory}
\vspace{-5pt}
For $T$ average CG iterations per right-hand side, total curvature computation costs $O((n+m)K(T+1))$. Under balanced workloads, ideal parallel time is $O\!\left(\frac{(n+m)K(T+1)}P+n+m\right)+T_{\mathrm{comm}}$, where $T_{\mathrm{comm}}$ accounts for data distribution and result aggregation.

\setlength{\textfloatsep}{8pt plus 1pt minus 2pt}

\begin{table*}[htbp]
\centering
\caption{\small Node classification results (\%) of different sampling methods. The best results are marked in bold.}
\footnotesize
\begin{tabular}{lccccccc}
\toprule
Method & PubMed & Amazon Photo & Coauthor CS & Flickr & ogbn-arxiv & Reddit & ogbn-products \\
\midrule
ClusterGCN & 87.91$\pm$0.41 & 93.40$\pm$0.71 & 93.36$\pm$0.20 & 51.57$\pm$0.41 & 64.64$\pm$0.57 & 91.36$\pm$0.29 & 72.44$\pm$1.16 \\
GraphSAGE & 88.03$\pm$0.48 & 89.86$\pm$0.74 & 93.14$\pm$0.25 & 44.18$\pm$2.44 & 56.95$\pm$1.80 & 90.50$\pm$ 0.10& 69.17$\pm$0.92 \\
FastGCN & 85.83$\pm$0.57 & 92.05$\pm$0.72 & 92.17$\pm$0.36 & 42.21$\pm$0.35 & 58.92$\pm$1.42 & 86.50$\pm$0.77 & 70.82$\pm$0.36 \\
GraphSAINT(node) & 88.01$\pm$0.09 & 93.06$\pm$0.13 & 93.80$\pm$0.15 & 51.31$\pm$0.16 & 65.60$\pm$0.19 & 87.12$\pm$0.30 & 77.68$\pm$0.32 \\
GraphSAINT(edge) & 87.70$\pm$0.02 & 93.49$\pm$0.20 & 93.71$\pm$0.09 & 51.25$\pm$0.21 & 62.27$\pm$0.19 & 82.21$\pm$0.26 & 58.92$\pm$0.14 \\
REGNN & 87.85$\pm$0.32 & 92.81$\pm$0.91 & 91.36$\pm$0.12 & 51.68$\pm$0.13 & 65.87$\pm$0.25 & \textbf{94.71$\pm$0.30} & 72.40$\pm$0.02 \\
LC-sub & 88.08$\pm$0.04 & 93.70$\pm$0.18 & 93.97$\pm$0.20 & 51.18$\pm$0.10 & 64.94$\pm$0.33 & 87.50$\pm$0.24 & 77.67$\pm$0.12 \\
ORG-sub & 87.96$\pm$0.22 & 93.76$\pm$0.09 & 93.79$\pm$0.05 & 51.27$\pm$0.20 & 63.68$\pm$0.61 & 89.50$\pm$0.22 & 76.76$\pm$0.11 \\
\midrule
LC-sub(ERC-LG) & 87.97$\pm$0.15 & 93.61$\pm$0.23 & 93.92$\pm$0.05 & 51.18$\pm$0.14 & 64.82$\pm$0.33 & 90.47$\pm$0.09 & 77.42$\pm$0.17 \\
ORG-sub(ERC-LG) & \textbf{88.12$\pm$0.09} & \textbf{93.83$\pm$0.30} & 93.81$\pm$0.14 & 51.27$\pm$0.12 & 61.12$\pm$0.21 & 91.67$\pm$0.04 & 76.20$\pm$0.30 \\
ERC-LG-sub(node) & 87.94$\pm$0.10 & 93.24$\pm$0.27 & 93.90$\pm$0.23 & 51.32$\pm$0.11 & 65.60$\pm$0.15 & 87.41$\pm$0.23 & 77.50$\pm$0.24 \\
ERC-LG-sub(edge) & 88.07$\pm$0.10 & 93.63$\pm$0.22 & \textbf{94.05$\pm$0.03} & \textbf{52.06$\pm$0.18} & \textbf{66.67$\pm$0.22} & 90.35$\pm$0.18 & \textbf{77.78$\pm$0.35} \\
\bottomrule
\end{tabular}
\label{tab:node_classification}
\end{table*}

Streaming projection generation and distance accumulation limit peak memory per GPU to $O(n+m+nb)$, including sparse graph operators, edge accumulators, and CG workspace. This excludes downstream GNN features and activations. Each GPU retains an $O(n+m)$ graph copy; increasing $P$ distributes computation but does not reduce this storage requirement. Constructing the curvature-based sampling weights takes $O(n+m)$ time.

\vspace{-5pt}
\section{Experiments}
\vspace{-5pt}
Our experiments address three questions:
\textbf{RQ1}. How does the proposed curvature-guided sampling framework perform on large-scale graphs for downstream node classification?
\textbf{RQ2}. How closely does approximate ERC agree with the pseudoinverse-based reference in numerical values and signs? 
\textbf{RQ3}. How efficient is ERC-LG in runtime and GPU memory, and how does its runtime scale with the number of GPUs? Additional experimental results, parameter sensitivity analyses (e.g., $K$, $b$, and $\varepsilon$), and implementation details, together with the ERC-LG code, are publicly available at \url{https://github.com/cqfei/ERC-large-graph}.

\vspace{-18pt}
\subsection{Experimental Settings}
\vspace{-5pt}
The datasets used in our experiments consist of medium-scale datasets including PubMed\footnote{https://github.com/shchur/gnn-benchmark/tree/master/data/planetoid}, Amazon Photo\footnote{https://github.com/shchur/gnn-benchmark/tree/master/data/npz}, and Coauthor CS\footnotemark[2], as well as large-scale datasets Flickr\footnote{https://github.com/GraphSAINT/GraphSAINT}, ogbn-arxiv\footnote{https://github.com/snap-stanford/ogb}, Reddit\footnote{https://snap.stanford.edu/graphsage/}, and ogbn-products\footnotemark[4].
We select representative large-graph GNN methods including ClusterGCN~\cite{chiang2019clustergcn}, GraphSAGE~\cite{hamilton2017graphsage}, FastGCN~\cite{chen2018fastgcn}, GraphSAINT~\cite{zeng2020graphsaint}, and REGNN~\cite{dingandwei2025}, as well as the curvature-aware large-graph sampling methods ORG-sub~\cite{wu2023subsampling} and LC-sub~\cite{shu2023109475} as our baselines.
To compare the impact of sampling quality on downstream tasks, we adopt the same two-layer GCN as the downstream model for all methods, following the training protocol of GraphSAINT for 200 epochs. In each epoch, a fixed number of subgraphs are sampled, and mini-batch training is performed on these sampled subgraphs.
For PubMed, ogbn-arxiv, and ogbn-products, we follow the official dataset splits, while 10-fold cross-validation is adopted for the remaining datasets. Performance is evaluated by the mean classification accuracy and standard deviation. All the models are implemented in PyTorch 1.12.0 and Python 3.9. Experiments are conducted on a server equipped with an Intel(R) Xeon(R) Gold 6248R CPU, 8 × NVIDIA RTX 3090 GPUs (24GB VRAM), and 256GB RAM. For ORC\footnote{https://github.com/saibalmars/GraphRicciCurvature} and ERC-REG\footnote{https://github.com/cqfei/resistance-curvature}, we adopt their public Python implementations, respectively.

\vspace{-5pt}
\subsection{Experimental Results}
\vspace{-5pt}

\subsubsection{Node Classification on Sampled Subgraphs}
\vspace{-5pt}
We evaluate four ERC-LG-based sampling variants in two groups. LC-sub(ERC-LG) and ORG-sub(ERC-LG) retain their original
sampling frameworks but replace the curvature
estimators with ERC-LG.
ERC-LG-sub(node) and ERC-LG-sub(edge) follow GraphSAINT
with sampling probabilities proportional to
the curvature-based weights in
Eqs.~\eqref{eq:node_probability}
and~\eqref{eq:edge_probability}, respectively.


Table~\ref{tab:node_classification} shows that
ERC-LG-based variants achieve the highest mean
accuracy on six of seven datasets, while REGNN
leads on Reddit.
ERC-LG-sub(edge) ranks first on four datasets and
outperforms GraphSAINT(edge) on all seven;
ERC-LG-sub(node) performs similarly to GraphSAINT(node).
LC-sub(ERC-LG) remains close to LC-sub on six datasets
and improves Reddit accuracy.
ORG-sub(ERC-LG) leads on PubMed and Amazon Photo
but underperforms ORG-sub on ogbn-arxiv and
ogbn-products, indicating that the benefits of
curvature substitution depend on the dataset.

\vspace{-15pt}
\subsubsection{Numerical Agreement of Curvature Estimates}
\vspace{-5pt}

We evaluate numerical agreement by comparing ERC-REG (diagonal perturbation), ERC-JL (JL projection), and ERC-LG against pseudoinverse-based ERC on stochastic block model (SBM).
Metrics include mean absolute error (MAE),
Spearman correlation, and sign agreement (Sign),
with $K$ denoting the projection dimension.
Table~\ref{tab:curvature_estimation} shows that
ERC-REG achieves an MAE of $10^{-5}$.
Increasing $K$ improves ERC-LG's agreement,
reaching an MAE of 0.00937, Spearman correlation
of 0.9687, and Sign of 92.39\% at $K=4096$.
ERC-LG and ERC-JL yield nearly identical metrics
across all tested dimensions.
Thus, incorporating CG yields comparable agreement
with the reference, while larger projection
dimensions improve approximation accuracy.

\vspace{-5pt}
\begin{table}[htbp]
\centering
\caption{\small Curvature approximation results on SBM dataset}
\footnotesize
\begin{tabular}{clccc}
\toprule
$K$ & Method & MAE & Spearman & Sign \\
\midrule
- & ERC-REG & 0.00001 & 1.0000 & 1.0000 \\
1024 & ERC-JL & 0.01793 & 0.8900 & 0.8555 \\
     & ERC-LG   & 0.01794 & 0.8900 & 0.8555 \\
2048 & ERC-JL & 0.01303 & 0.9401 & 0.8943 \\
     & ERC-LG   & 0.01304 & 0.9401 & 0.8943 \\
4096 & ERC-JL & 0.00937 & 0.9687 & 0.9239 \\
     & ERC-LG   & 0.00937 & 0.9687 & 0.9239 \\
\bottomrule
\end{tabular}
\label{tab:curvature_estimation}
\vspace{-5pt}
\end{table}

\vspace{-5pt}
\subsubsection{Computational Efficiency and GPU Memory Usage}
\vspace{-5pt}

\begin{figure}[t]
\vspace{-5pt}
\centering
\begin{minipage}[t]{0.48\linewidth}
    \vspace{0pt}
    \centering
   
    \captionof{table}{\scriptsize GPU memory and runtime}
    \label{tab:mem_runtime}

    \footnotesize
    \setlength{\tabcolsep}{1pt}
    \renewcommand{\arraystretch}{1.12}

    \begin{tabular*}{\linewidth}
        {@{\extracolsep{\fill}}lccc@{}}
        \toprule
        Setting & Flickr & ogbn-arxiv & Reddit \\
        \midrule
        $b=256$ & 2950 & 4924 & 10118 \\
        $b=512$ & 4708 & 8646 & 14689 \\
        \midrule
        ERC-LG & 27 & 131 & 1382 \\
        ERC-CG & 253 & 4518 & - \\
        \bottomrule
    \end{tabular*}

    \par
    \smallskip
    {\scriptsize
    Upper: memory (MB), $K=2048$; lower: runtime (s), $P=2$, $b=256$. ``-'' denotes a timeout. \par}
\end{minipage}
\hfill
\begin{minipage}[t]{0.5\linewidth}
    \vspace{0pt}
    \centering
    \includegraphics[width=\linewidth]
    {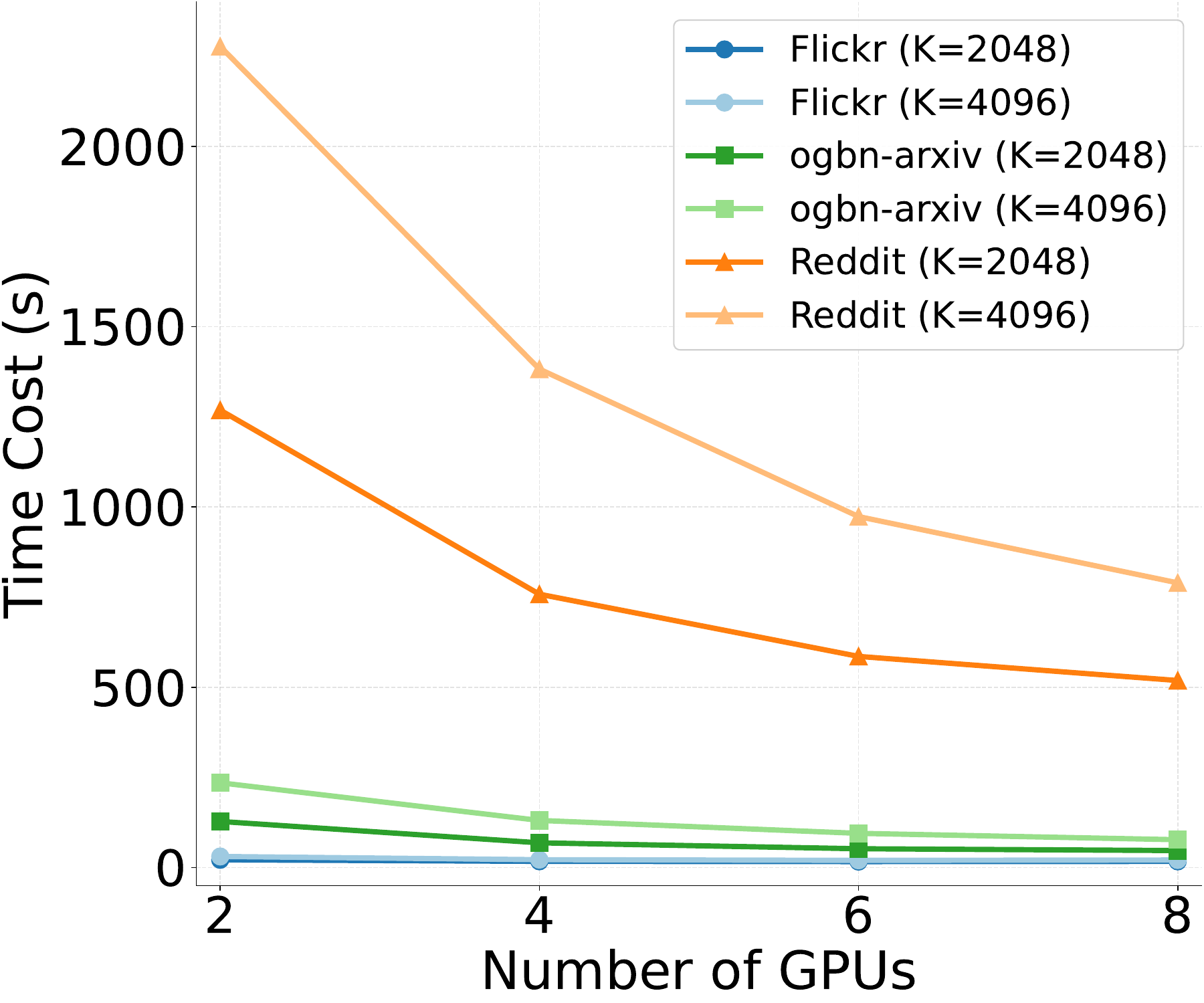}
\vspace{-25pt}
    \caption{\scriptsize Multi-GPU curvature runtime}
    \label{fig:gpu_runtime}
\end{minipage}
\end{figure}
Table~\ref{tab:mem_runtime} summarizes peak memory usage per GPU (upper block) and runtime (lower block). Reducing the batch size from 512 to 256 lowers memory usage on all three datasets, including a reduction from 14,689 to 10,118 MB on Reddit.

Compared with ERC-CG, which uses CG without JL projection, ERC-LG is approximately $9.4\times$ faster on Flickr and $34.5\times$ faster on ogbn-arxiv. It also completes curvature computation on Reddit in 1382 seconds, supporting the computational benefit of JL projection. On ogbn-products, the maximum feasible batch size in our implementation is \(b=76\) due to the 24 GB GPU memory limit. ERC-LG nevertheless completes curvature computation in 11366s on 4 GPUs ($K=2048$), demonstrating its applicability to graphs beyond the sizes reported in Table~\ref{tab:mem_runtime}.

Figure~\ref{fig:gpu_runtime} shows that runtime decreases as the GPU count increases from 2 to 8 on all three datasets, for both $K=2048$ and $K=4096$. Multi-GPU execution therefore further reduces curvature computation time.

\vspace{-5pt}
\section{Conclusion}
\vspace{-5pt}
We proposed ERC-LG, combining JL projection and regularized multi-GPU batched CG to approximate ERC without explicit pseudoinverse computation. The resulting curvature guides node and edge sampling for GNN training. Experiments show its numerical accuracy and computational efficiency, with ERC-LG-based sampling variants achieving the highest mean classification accuracy on six of seven datasets. These results support ERC-LG as a practical means of incorporating resistance geometry into large-scale graph sampling and representation learning.
\section{Acknowledgments}
This work was supported in part by the National Natural Science Foundation of China (No. 62406315), China Postdoctoral Science Foundation (2025M771504), Basic and Applied Basic Research Foundation of Guangdong Province (2024A1515110108),  Key Research and Development Projects of Shaanxi Province (2025SF-YBXM-023), and Shaanxi Provincial Public Health Scientific Research Innovation Team Project (202511).

\bibliographystyle{IEEEbib}
\bibliography{strings,refs}

\end{document}